\documentclass[sigconf, final]{acmart}

\usepackage{booktabs} 
\usepackage{longtable}

\usepackage{subcaption}

\usepackage{comment}            
\usepackage{amsmath}
\usepackage{booktabs}
\usepackage{graphicx}
\usepackage{amsmath}
\usepackage{float}
\usepackage{multirow}
\usepackage[load-configurations=version-1]{siunitx} 
\newcommand{\RNum}[1]{%
  \textup{\uppercase\expandafter{\romannumeral#1}}%
}

\newcommand\given[1][]{\:#1\vert\:}

\setcopyright{rightsretained}

\acmConference

\copyrightyear{2018} 
\acmYear{2018} 
\setcopyright{acmlicensed}
\acmConference[ICISDM '18]{2018 2nd International Conference on Information System and Data Mining ICISDM '18}{April 9--11, 2018}{Lakeland, FL, USA}
\acmBooktitle{ICISDM '18: 2018 2nd International Conference on Information System and Data Mining ICISDM '18, April 9--11, 2018, Lakeland, FL, USA}
\acmPrice{15.00}
\acmDOI{10.1145/3206098.3206111}
\acmISBN{978-1-4503-6354-9/18/04}

\begin{document}
\title{RMDL: \underline{R}andom \underline{M}ultimodel \underline{D}eep \underline{L}earning for Classification}

\author{Kamran Kowsari}
\orcid{1234-5678-9012}
\authornote{Sensing Systems for Health Lab, University of Virginia, Charlottesville, VA USA}
\affiliation{%
  \institution{Department of System and Information Engineering,\\
 University of Virginia}
  \streetaddress{P.O. Box 1212}
  \city{Charlottesville}
  \state{VA, USA}
  \postcode{22903}
}
\email{kk7nc@virginia.edu}

\author{Mojtaba Heidarysafa}
\affiliation{%
  \institution{Department of System and Information Engineering,\\
University of Virginia}
  \streetaddress{P.O. Box 1212}
  \city{Charlottesville}
  \state{VA, USA}
  \postcode{22903}
}
\email{mh4pk@virginia.edu}

\author{Donald E. Brown}
\authornote{\label{note1}Data Science Institute, University of Virginia
Charlottesville, VA USA}
\authornote{Predictive Technology Laboratory, University of Virginia, Charlottesville, VA USA}
\affiliation{%
  \institution{Department of System and Information Engineering,\\
University of Virginia}
  \streetaddress{P.O. Box 1212}
  \city{Charlottesville}
  \state{VA, USA}
  \postcode{22903}}
\email{deb@virginia.edu}

\author{Kiana Jafari Meimandi}
\affiliation{%
  \institution{Department of System and Information Engineering,\\
University of Virginia}
  \streetaddress{P.O. Box 1212}
  \city{Charlottesville}
  \state{VA, USA}
  \postcode{22903}}
\email{kj6vd@virginia.edu}

\author{Laura E. Barnes}
\authornotemark[1]
\authornotemark[2]

\affiliation{%
  \institution{Department of System and Information Engineering,\\
University of Virginia}
  \streetaddress{P.O. Box 1212}
  \city{Charlottesville}
  \state{VA, USA}
  \postcode{22903}}
\email{lb3dp@virginia.edu}

\renewcommand{\shortauthors}{K. Kowsari et al.}

\begin{abstract}
The continually increasing number of complex datasets each year necessitates ever improving machine learning methods for robust and accurate categorization of these data. This paper introduces Random Multimodel Deep Learning~(RMDL): a new ensemble, deep learning approach for classification. Deep learning models have achieved state-of-the-art results across many domains.  RMDL solves the problem of finding the best deep learning structure and architecture while simultaneously improving robustness and accuracy through ensembles of deep learning architectures. RDML can accept as input a variety data to include text, video, images, and symbolic. This paper describes RMDL and shows test results for image and text data including MNIST, CIFAR-10, WOS, Reuters, IMDB, and 20newsgroup. These test results show that RDML produces consistently better performance than standard methods over a broad range of data types and classification problems.\footnote{Code is shared as an open source tool at \url{https://github.com/kk7nc/RMDL}}
\end{abstract}

%
%



\maketitle

\section{Introduction}\label{sec:introduction}
Categorization and classification with complex data such as images, documents, and video are central challenges in the data science community. Recently, there has been an increasing body of work using deep learning structures and architectures for such problems. 
However, the majority of these deep architectures are designed for a specific type of data or domain. There is a need to develop more general information processing methods for classification and categorization across a broad range of data types. 

While many researchers have successfully used deep learning for  classification problems (\textit{e.g.,} see~\cite{kowsari2017HDLTex,lecun2015deep,lee2009convolutional,chung2014empirical,turan2017deep}), 
the central problem remains as to which deep learning architecture~(DNN, CNN, or RNN) and structure~(how many nodes~(units) and hidden layers) is more efficient for different types of data and applications. The favored approach to this problem is 
trial and error for the specific application and dataset.

This paper describes an approach to this challenge using ensembles of deep learning architectures. This approach, called Random Multimodel Deep Learning~(RMDL), 
uses three different deep learning architectures: Deep Neural Networks~(DNN), Convolutional Neural Netwroks~(CNN), and Recurrent Neural Networks~(RNN). Test results with a variety of data types demonstrate that this new approach is highly accurate, robust and efficient.

The three basic deep learning architectures use different feature space methods as input layers. 
For instance, for feature extraction from text, DNN 
uses term frequency-inverse document frequency~(TF-IDF)~\cite{robertson2004understanding}. RDML searches across randomly generated hyperparameters for the number of hidden layers and nodes~(desity) in each hidden layer in the DNN.
CNN has been well designed for image classification. RMDL finds choices for hyperparameters in CNN using random feature maps and random numbers of hidden layers. CNN can be used for more than image data. The structures for CNN used by RMDL are 1D convolutional layer for text, 2D for images and 3D for video processings.
RNN architectures are used primarily for text classification. RMDL uses two specific RNN structures: Gated Recurrent Units~(GRUs) and Long Short-Term Memory~(LSTM). The number of GRU or LSTM units and hidden layers used by the RDML are also the results of search over randomly generated hyperparameters.

The main contributions of this work are as follows:~\RNum{1}) Description of an ensemble approach to deep learning which makes the final model more robust and accurate.~\RNum{2}) Use of different optimization techniques in training the models to stabilize the classification task.~\RNum{3}) Different feature extraction approaches for each Random Deep Leaning~(RDL) model in order to better understand the feature space~(specially for text and video data).~\RNum{4}) Use of dropout in each individual RDL to address over-fitting.~\RNum{5}) Use of majority voting among the ~$n$ RDL models. This majority vote from the ensemble of RDL models improves the accuracy and robustness of results. Specifically, if~$k$ number of RDL models produce inaccuracies or overfit classifications and $n > k$, the overall system is robust and accurate~\RNum{6}) Finally, the RMDL has ability to process a variety of data types such as text, images and videos.

The rest of this paper is organized as follows: Section~\ref{sec:related} gives related work for feature extraction, other classification techniques, and deep learning for classification task;   Section~\ref{sec:baseline} describes current techniques for classification tasks which are used as our baseline; Section~\ref{sec:method} describes Random Multimodel Deep Learning methods and the architecture for RMDL including Section~\ref{subsect:Feature} shows feature extraction in RMDL, Section~\ref{subsect:RMDL} talks about overall view of RMDL; Section~\ref{subsec:Deep_learning} addresses the deep learning structure used in this model, Section~\ref{subsec:Optimization} discusses optimization problem; Section~\ref{subsec:Evaluation} talks about evaluation of these techniques; Section~\ref{sec:results} shows the experimental results which includes the accuracy and performance of RMDL; and finally, Section~\ref{sec:Conclusion} presents discussion and conclusions of our work.

\vspace{-0.1in}
\section{Related Work}\label{sec:related}
Researchers from a variety of disciplines have produced work relevant to the approach described in this paper. We have organized this work into three areas: 
~\RNum{1})~Feature extraction; ~\RNum{2})~Classification methods and techniques~(baseline and other related methods); and ~\RNum{3})~Deep learning for classification.

\textbf{Feature Extraction:}\label{subsec:related1}
Feature extraction is a significant part of machine learning especially for text, image, and video data. 
 Text and many biomedical datasets are mostly unstructured data from which we need to generate a meaningful and structures for use by machine learning algorithms. As an early example, L. Krueger~\textit{et. al.} in~1979~\cite{krueger1979letter} introduced an effective method for feature extraction for text categorization. This feature extraction method is based on word counting to create a structure for statistical learning. Even earlier work by H. Luhn~\cite{luhn1957statistical} introduced weighted values for each word and then G. Salton~\textit{et. al.} in 1988~\cite{salton1988term} modified the weights of words by frequency counts called term frequency-inverse document frequency~(TF-IDF). The TF-IDF vectors measure the number of times a word appears in the document weighted by the inverse frequency of the commonality of the word across documents. Although, the TF-IDF and word counting are simple and intuitive feature extraction methods, they do not capture relationships between words as sequences. Recently, T. Mikolov~\textit{et. al.}~\cite{mikolov2013efficient} introduced an improved technique for feature extraction from text using the concept of embedding or placing the word into a vector space based on context.  This approach to word embedding, called \textit{Word2Vec}, solves the problem of 
 representing contextual word relationships in a computable feature space. Building on these ideas, J. Pennington~\textit{et. al.} in~2014~\cite{pennington2014glove} developed a learning vector space representation of the words called~\textit{Glove} and deployed it in Stanford NLP lab. The RMDL approach described in this paper uses Glove for feature extraction from textual data.

\textbf{Classification Methods and Techniques:}\label{subsec:related2} 
Over the last 50 years, many supervised learning classification techniques have been developed and implemented in software to accurately label  data. For example, the researchers, K. Murphy in~2006~\cite{murphy2006naive} and I. Rish in~2001~\cite{rish2001empirical} introduced the Na\"ive Bayes Classifier~(NBC) as a simple approach to the more general respresentation of the supervised learning classification problem. This approach has provided a useful technique for text classification and information retrieval applications. As with most supervised learning classification techniques, NBC takes an input vector of numeric or categorical data values and produce the probability for each possible output labels. This approach is fast and efficient for text classification, but NBC has important limitations. Namely, the order of the sequences in text is not reflected on the output probability because for text analysis, na\"ive bayes uses a bag of words approach for feature extraction. Because of its popularity, this paper uses NBC as one of the baseline methods for comparison with RMDL. Another  popular classification technique is Support Vector Machines~(SVM), which has proven quite accurate over a wide variety of data. This technique constructs a set of hyper-planes in a transformed feature space. This transformation is not performed explicitly but rather through the kernal trick which allows the SVM classifier to perform well with highly nonlinear relationships between the predictor and response variables in the data.  A variety of approaches have been developed to further extend the basic methodology and obtain greater accuracy. C. Yu~\textit{et. al.} in 2009~\cite{yu2009learning} introduced  latent variables into the discriminative model as a new structure for SVM, and S. Tong~\textit{et. al.} in 2001~\cite{tong2001support} added active learning using SVM for text classification. For a large volume of data and  datasets with a huge number of features~(such as text), SVM implementations are computationally complex. Another technique that helps mediate the computational complexity of the SVM for classification tasks is stochastic gradient descent classifier~(SGDClassifier)~\cite{kabir2015bangla} which has been widely used in both text and image classification. SGDClassifier is an iterative model for large datasets. The model is trained based on the SGD optimizer iteratively.

\textbf{Deep Learning:}\label{subsec:related3}
Neural networks derive their architecture as a relatively simply representation of the neurons in the human's brain. They are essentially weighte combinations of inputs the pass through multiple non-linear functions. Neural networks use an iterative learning method known as back-propagation and an optimizer~(such as stochastic gradient descent~(SGD)). 

Deep Neural Networks~(DNN) are based on simple neural networks architectures but they contain multiple hidden layers. These networks have been widely used for classification. For example, D. Cire{\c{s}}An~\textit{et. al.} in 2012~\cite{ci2012multitraffic} used multi-column deep neural networks for classification tasks, where  multi-column deep neural networks use DNN architectures. Convolutional Neural Networks~(CNN) provide a different architectural approach to learning with neural networks. The main idea of CNN is to use feed-forward networks with convolutional layers that include local and global pooling layers. A. Krizhevsky in 2012~\cite{krizhevsky2012imagenet} used CNN, but they have used~$2D$ convolutional layers combined with the ~$2D$ feature space of the image. Another example of CNN in~\cite{lecun2015deep} showed excellent accuracy for image classification.  This architecture can also be used for text classification as shown in the work of \cite{kim2014convolutional}. For text and sequences,~$1D$ convolutional layers are used with word embeddings as the input feature space. The final type of deep learning architecture is Recurrent Neural Networks~(RNN) where outputs from the neurons are fed back into the network as inputs for the next step. Some recent extensions to this architecture uses Gated Recurrent Units~(GRUs)~\cite{chung2014empirical} or Long Short-Term Memory~(LSTM) units~\cite{hochreiter1997long}. These new units help control for instability problems in the original network architecure. RNN have been successfully used for natural language processing~\cite{mikolov2010recurrent}. Recently, Z. Yang~\textit{et. al.} in~2016~\cite{yang2016hierarchical} developed hierarchical attention networks for document classification. These networks have two important characteristics: hierarchical structure and an attention mechanism at word and sentence level. 

New work has combined these three basic models of the deep learning structure and developed a novel technique for enhancing accuracy and robustness. The work of M. Turan~\textit{et. al.} in 2017~\cite{turan2017deep} and M. Liang~\textit{et. al.}in 2015~\cite{liang2015recurrent} implemented innovative combinations of CNN and RNN called \textit{A Recurrent Convolutional Neural Network~(RCNN)}. K. Kowsari~\textit{et. al.} in 2017~ ~\cite{kowsari2017HDLTex} introduced hierarchical deep learning for text classification~(HDLTex) which is a combination of all deep learning techniques in a hierarchical structure for document classification has improved accuracy over traditional methods. The work in this paper builds on these ideas, spcifically the work of~\cite{kowsari2017HDLTex} to provide a more general approach to supervised learning for classification.

\section{Baseline}\label{sec:baseline}
In this paper, we use both contemporary and traditional techniques of document and image classification as our baselines. The baselines of image and text classification are different due to feature extraction and structure of model; thus, text and image classification's baselines are described separately in the following section.
\subsection{Text Classification Baselines}
Text classification techniques which are used as our baselines to evaluate our model are as follows: regular deep models such as Recurrent Neural Networks~(RNN), Convolutional Neural Networks~(CNN), and Deep Neural Networks~(DNN). Also, we have used two different techniques of Support Vector Machine~(SVM), na\"{i}ve bayes classification~(NBC), and finally Hierarchical Deep Learning for Text Classification~(HDLTex)~\cite{kowsari2017HDLTex}.
\subsubsection{Deep Learning}
The baseline, we used in this paper is Deep Learning without hierarchical levels. An example of hierarchical levels' structure is~\cite{yang2016hierarchical} that has been used as one of our baselines for text classification. In our methods' Section~\ref{method}, we will explain the basic models of deep learning such as DNN, CNN, and RNN which are used as part of RMDL model. 
\subsubsection{Support Vector Machine~(SVM)}\label{me_svm}
The original version of SVM was introduced by  Vapnik, VN and Chervonenkis, A Ya~\cite{chervonenkis2013early} in 1963. The early 1990s, nonlinear version was addressed in~\cite{boser1992training}.
\subsubsection*{Multi-class SVM}
The original version of SVM is used for binary classification, so for multi class we need to generate Multimodel or MSVM. One-Vs-One is a technique for multi-class SVM and needs to build N(N-1) classifiers.

The natural way to solve k-class problem is to construct a decision function of all $k$ classes at once~\cite{chen2016turning,weston1998multi}.  
Another technique of multi-class classification using SVM is All-against-One. In SVM, many different methods are available for feature extraction such as word sequences feature extracting~\cite{zhang2008text}, and Term frequency-inverse document frequency~(TF-IDF).
\subsubsection*{String Kernel}\label{String_kernel}
The basic idea of String Kernel~(SK) is using~$\Phi(.)$ for mapping string in the feature space; therefore, the only different between the three techniques are the way they map the string into feature space. For many applications such as text, DNA, and protein classification, Spectrum Kernel~(SP) is addressed~\cite{leslie2002spectrum,eskin2002mismatch}. The basic idea of SP is counting number of time a word appears in string $x_i$  as feature map where defining feature maps from $x \rightarrow \mathbb{R}^{l^k}$

 Mismatch Kernel is the other stable way to map the string into feature space. The key idea is using $k$ which stands for $k-mer$ or size of the word and allow to have~$m$ mismatch in feature space~\cite{leslie2004mismatch}. The main problem of SVM for string sequences is time complexity of these models. S. Ritambhara~\textit{et. al.} in~2017~\cite{singh2017gakco} addressed the problem of time for gap k-mers kernel called \textit{GaKCo} which is used only for protein and DNA sequences.

\subsubsection{Stacking Support Vector Machine~(SVM)}
Stacking SVMs is used as another baseline method for comparison with RMDL, but this technique is used only for hierarchical labeled datasets. The stacking SVM provides an ensemble of individual SVM classifiers and generally produces more accurate results than single-SVM models  ~\cite{sun2001hierarchical,sebastiani2002machine}.

\subsubsection{Na\"{i}ve Bayes Classification~(NBC)}\label{me_NB}
This technique has been used in industry and academia for a long time, and it is the most traditional method of text categorization which is widely used in Information Retrieval~\cite{manning2008introduction}. If the number of $n$ documents, fit into $k$ categories, the predicted class as output is $c \in C$. Na\"{i}ve bayes is a simple algorithm using na\"{i}ve bayes rule described  as follows:
\begin{equation}
P(c \given d) = \frac{P(d \given c)P(c)}{P(d)}
\end{equation}
where $d$ is document, $c$ indicates classes.
\begin{equation} \label{eq1}
\begin{split}
C_{MAP} &=~  arg \max_{c \in C} P(d \given c)P(c) \\&= arg \max_{c\in C} P(x_1,x_2,...,x_n \given c) p(c)
\end{split}
\end{equation}
The baseline of this paper is word level of NBC~\cite{kim2006some} as follows:
\begin{equation}
    P(c_j \given d_i;\hat{\theta}) = \frac{P(c_j \given \hat{\theta})P(d_i \given c_j ; \hat{\theta_j})}{P(d_i \given \hat{\theta})}
\end{equation}

\subsubsection{Hierarchical Deep Learning for Text Classification~(HDLTex)} This technique is used as one of our baselines for hierarchical labeled datasets. When documents are organized hierarchically, multi-class approaches are difficult to apply using traditional supervised learning methods. The HDLTex~\cite{kowsari2017HDLTex} introduced a new approach to hierarchical document classification that combines multiple deep learning approaches to produce hierarchical classification. The primary contribution of HDLTex research is hierarchical classification of documents.  A traditional multi-class classification technique can work well for a limited number of classes, but performance drops with increasing number of classes, as is present in hierarchically organized documents. HDLTex solved this problem by creating architectures that specialize deep learning approaches for their level of the document hierarchy.

\vspace{-0.1in}
\subsection{Image Classification Baselines}\label{subsec:baseline_image}
For image classification, we have five baselines as follows: Deep L2-SVM~\cite{tang2013deep}, Maxout Network~\cite{goodfellow2013maxout}, BinaryConnect~\cite{courbariaux2015binaryconnect}, PCANet-1 ~\cite{chan2015pcanet}, and gcForest~\cite{zhou2017deep}.\\
\textit{Deep L2-SVM:} This technique is known as deep learning using linear support vector machines which simply softmax is replaced with linear SVMs~\cite{tang2013deep}.
\\ \textit{Maxout Network:} I. Goodfellow~\textit{et. al.}~in~2013~\cite{goodfellow2013maxout} defined a simple novel model called \textit{maxout} (named because its outputs' layer is a set of max of inputs' layer, and it is a natural companion to dropout). Their design both facilitates optimization by using dropout, and also improves the accuracy of dropout's model.\\
\textit{BinaryConnect:} M. Courbariaux~\textit{et. al.}~in~2015~\cite{courbariaux2015binaryconnect} worked on training Deep Neural Networks~(DNN) with binary weights during propagations. They have introduced a binarization scheme for binary weights during forward and backward propagations~(\textup{BinaryConnect}) which is mainly used for image classification. BinaryConnect is used as our baseline for RMDL on image classification.\\ \textit{PCANet:} I. Chan~\textit{et. al.}~in~2015~\cite{chan2015pcanet} is simple way of deep learning for image classification which uses~CNN structure. Their technique is one of the basic and efficient methods of deep learning. The CNN structure they've used, is part of RMDL with significant differences that they use: \RNum{1}) cascaded principal component analysis (PCA); \RNum{2}) binary hashing; and  \RNum{3}) blockwise histograms, and also number of hidden layers and nodes in RMDL is selected automatically.\\ \textit{gcForest~(Deep Forest):} Z. Zhou~\textit{et. al.}~in~2017~\cite{zhou2017deep} introduced a decision tree ensemble approach with high performance as an alternative to deep neural networks. Deep forest creates multi level of forests as decision trees.

\section{Method}\label{method}\label{sec:method}

The novelty of this work is in using multi random deep learning models including DNN, RNN, and CNN techniques for text and image classification. The method section of this paper is organized as follows: first we describe RMDL and we discuss three techniques of deep learning architectures~(DNN, RNN, and CNN) which are trained in parallel. Next, we talk about multi optimizer techniques that are used in different random models. 
\subsection{Feature Extraction and Data Pre-processing}\label{subsect:Feature}
The feature extraction is divided into two main parts for RMDL~(Text and image). Text and sequential datasets are unstructured data, while the feature space is structured for image datasets.
\subsubsection{Image and 3D Object Feature Extraction}
Image features are the followings:~$h\times~w\times~c$ where~$h$ denotes the height of the image,~$w$ represents the width of image, and~$c$ is the color that has 3 dimensions~(RGB). For gray scale datasets such as $MNIST$ dataset, the feature space is~$h\times~w$. A 3D object in space contains~$n$ cloud points in space and each cloud point has~$6$ features which are~(\textit{x, y, z, R, G, and B}). The 3D object is unstructured due to number of cloud points since one object could be different with others. However, we could use simple instance down/up sampling to generate the structured datasets.
\begin{figure*}[t]
\centering
\includegraphics[width=0.65\textwidth]{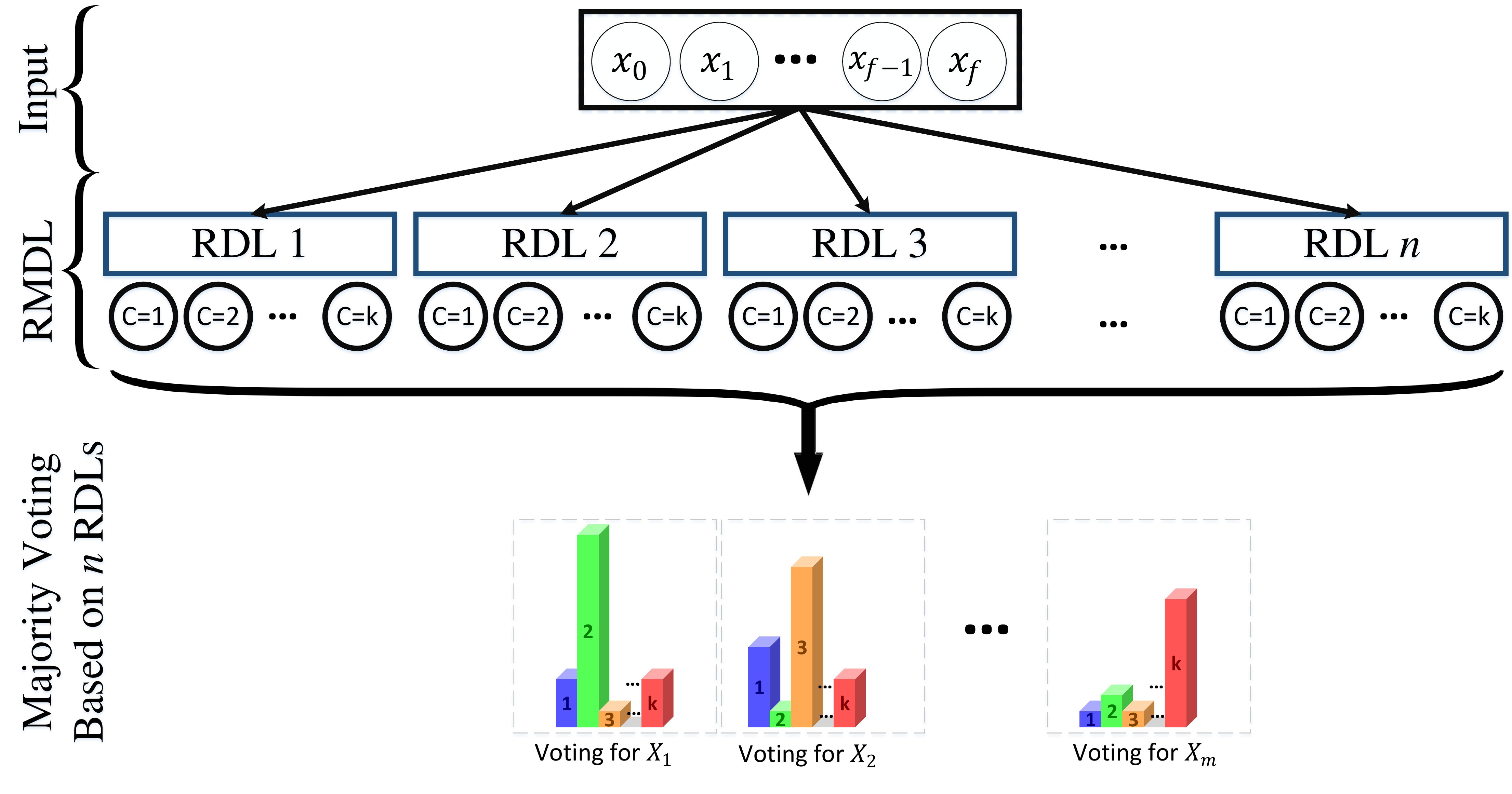}

\caption{Overview of RDML: \underline{R}andom \underline{M}ultimodel \underline{D}eep \underline{L}earning for classification that includes $n$ Random models which are $d$ random model of DNN classifiers, $c$ models of CNN classifiers, and $r$ RNN classifiers where~$r+c+d=n$.}\label{Fig_RMDL}
\vspace{-0.15in}

\end{figure*}

\subsubsection{Text and Sequences Feature Extraction}
In this paper we use several techniques of text feature extraction which are word embedding~(GloVe and Word2vec) and also TF-IDF. 
In this paper, we use word vectorization techniques~\cite{hotta2010word} for extracting features; Besides, we also can use N-gram representation as features for neural deep learning~\cite{kevselj2003n,dave2003mining}. For example, feature extraction in this model for the string "In this paper we introduced this technique" would be composed of the following:
\begin{itemize}
\item Feature count(1) \{ (In 1) , (this 2), (paper 1), (we 1), (introduced 1), (technique 1) \}

\item Feature count(2) \{ (In 1) , (this 2), (paper 1), (we 1), (introduced 1), (technique 1), (In this 1), (This Paper 1), ( paper we 1), ( we introduced 1), (introduced this 1), ( this technique 1) \}

\end{itemize}
Documents enter our models via features extracted from the text. We employed different feature extraction approaches for the deep learning architectures we built. For CNN and RNN, we used the text vector-space models using $200$ dimensions as described in GloVe~\cite{pennington2014glove}. A vector-space model is a mathematical mapping of the word space, defined as follows:
\begin{equation}
    d_j = (w_{1,j},w_{2,j},...,w_{i,j}...,w_{l_j,j})
\end{equation}
where $l_j$ is the length of the document $j$, and $w_{i,j}$ is the GloVe word embedding vectorization of word $i$ in document $j$.

\subsection{Random Multimodel Deep Learning}\label{subsect:RMDL}
Random Multimodel Deep Learning is a novel technique that we can use in any kind of dataset for classification. An overview of this technique is shown in Figure~\ref{Fig_RMDL} which contains multi Deep Neural Networks~(DNN), Deep Convolutional Neural Networks~(CNN), and Deep Recurrent Neural Networks~(RNN). The number of layers and nodes for all of these Deep learning multi models are generated randomly~(\textit{e.g.}~9 Random Models in RMDL constructed of~$3$ CNNs,~$3$ RNNs, and~$3$ DNNs, all of them are unique due to randomly creation). 
\begin{align}
\label{eq:majority}
M(y_{i1},y_{i2},...,y_{in}) =& \bigg\lfloor \frac{1}{2}+ \frac{(\sum_{j=1}^n y_{ij}) - \frac{1}{2}}{n}\bigg\rfloor
\end{align}
Where $n$ is the number of random models, and $y_{ij}$ is the output prediction of model for data point $i$ in  model $j$~(Equation~\ref{eq:majority} is used for binary classification, $k\in\{0~ \text{or}~1\}$). Output space uses majority vote for final $\hat{y_i}$. Therefore,~$\hat{y_i}$ is given as follows:
\begin{equation}
\hat{y_i} =  
\begin{bmatrix}
\hat{y}_{i1} ~
\hdots ~
\hat{y}_{ij}~
\hdots~
\hat{y}_{in}~
\end{bmatrix}^T
\end{equation} 

Where $n$ is number of random model, and $\hat{y}_{ij}$ shows the prediction of label of document or data point of $D_i \in \{x_i,y_i\}$ for model $j$ and $\hat{y}_{i,j}$ is defined as follows:
\begin{equation}
\hat{y}_{i,j} = arg \max_{k} [ softmax(y_{i,j}^*)]
\end{equation}
After all RDL models~(RMDL) are trained, the final prediction is calculated using majority vote of these models.

\subsection{Deep Learning in RMDL}\label{subsec:Deep_learning}
The RMDL model structure~(section~\ref{subsect:RMDL}) includes three basic architectures of deep learning in parallel. We describe each individual model separately. The final model contains~$d$ random DNNs~(Section~\ref{subsubsec:DNN}),~$r$ RNNs~(Section~\ref{subsubsec:RNN}), and~$c$ CNNs models~(Section~\ref{subsubsec:CNN}).
\begin{figure*}[t]
\centering
\includegraphics[width=0.88\textwidth]{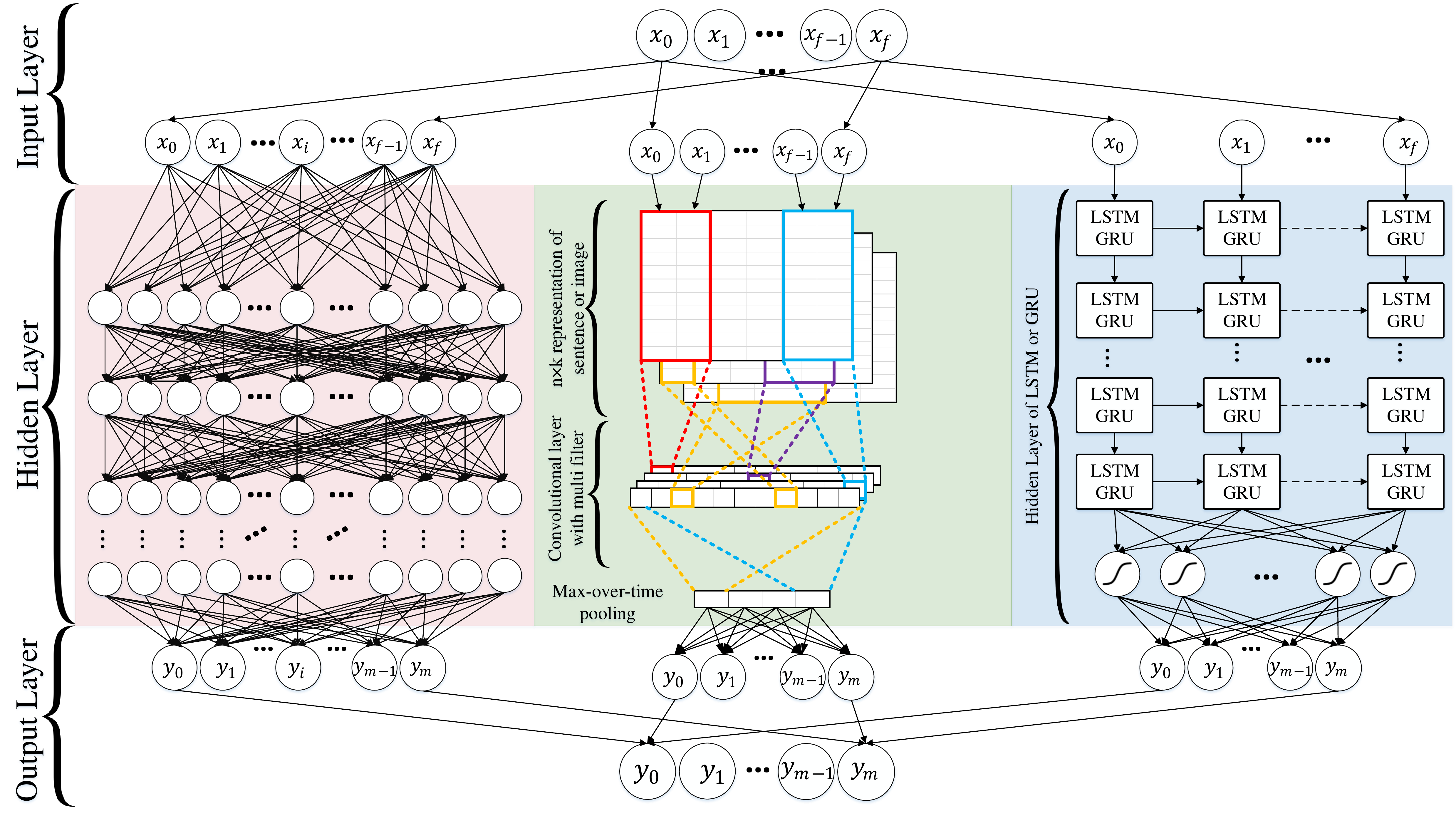}
\caption{\underline{R}andom \underline{M}ultimodel \underline{D}eep \underline{L}earning~(RDML) architecture for classification which includes~$3$ Random models, a DNN classifier at left, a Deep CNN classifier at middle, and a Deep RNN classifier at right~(each unit could be LSTM or GRU).}
\vspace{-0.15in}
\label{Fig_RMDL}

\end{figure*}
\subsubsection{Deep Neural Networks}\label{subsubsec:DNN}
Deep Neural Networks' structure is designed to learn by multi connection of layers that each layer only receives connection from previous and provides connections only to the next layer in hidden part. The input is a connection of feature space with first hidden layer for all random models. The output layer is number of classes for multi-class classification and only one output for binary classification. But our main contribution of this paper is that we have many training DNN for different purposes. In our techniques, we have multi-classes DNNs where each learning models is generated randomly~(number of nodes in each layer and also number of layers are completely random assigned). Our implementation of Deep Neural Networks~(DNN) is discriminative trained model that uses standard back-propagation algorithm using sigmoid~(equation~\ref{sigmoid}), ReLU~\cite{nair2010rectified}~(equation~\ref{relu}) as activation function. The output layer for multi-class classification, should use $Softmax$ equation~\ref{Softmax}.
\begin{align}
\label{sigmoid}
f(x) =& \frac{1}{1+e^{-x}}\in (0,1)\\
\label{relu}
f(x) =& \max(0,x)
\end{align}

\begin{align}
\label{Softmax}
\sigma(z)_j =& \frac{e^{z_j}}{\sum_{k=1}^K e^{z_k}}\\ 
\forall   ~j \in &\{1,\hdots, K\} \nonumber
\end{align}
 Given a set of example pairs $(x,y),x\in X,y\in Y$, the goal is to learn from these input and target space using hidden layers. In text classification, the input is string which is generated by vectorization of text. In Figure~\ref{Fig_RMDL} the left model shows how DNN contribute in RMDL.

\subsubsection{Recurrent Neural Networks~(RNN)}\label{subsubsec:RNN}
Another neural network architecture that contributes in RMDL is Recurrent Neural Networks~(RNN). RNN assigns more weights to the previous data points of sequence. Therefore, this technique is a powerful method for text, string and sequential data classification but also  could be used for image classification as we did in this work. In RNN the neural net considers the information of previous nodes in a very sophisticated method which allows for better semantic analysis of structures of dataset. General formulation of this concept is given in Equation~\ref{rnn_gen} where $x_t$ is the state at time $t$ and $\boldsymbol{u_t}$ refers to the input at step t.

\begin{equation}
\label{rnn_gen}
x_{t}=F(x_{t-1},\boldsymbol{u_t},\theta)
\end{equation}

More specifically, we can use weights to formulate the Equation~\ref{rnn_gen} with  specified parameters in Equation~\ref{rnn_spec}

\begin{equation}\label{rnn_spec}
x_{t}=\mathbf{W_{rec}}\sigma(x_{t-1})+\mathbf{W_{in}}\mathbf{u_t}+\mathbf{b}
\end{equation}
Where $\mathbf{W_{rec}}$ refers to recurrent matrix weight, $\mathbf{W_{in}}$ refers to input weights, $\mathbf{b}$ is the bias and $\sigma$ denotes an element-wise function. 

\begin{figure}[b]
\vspace{-0.15in}
\centering
\includegraphics[width=0.75\columnwidth]{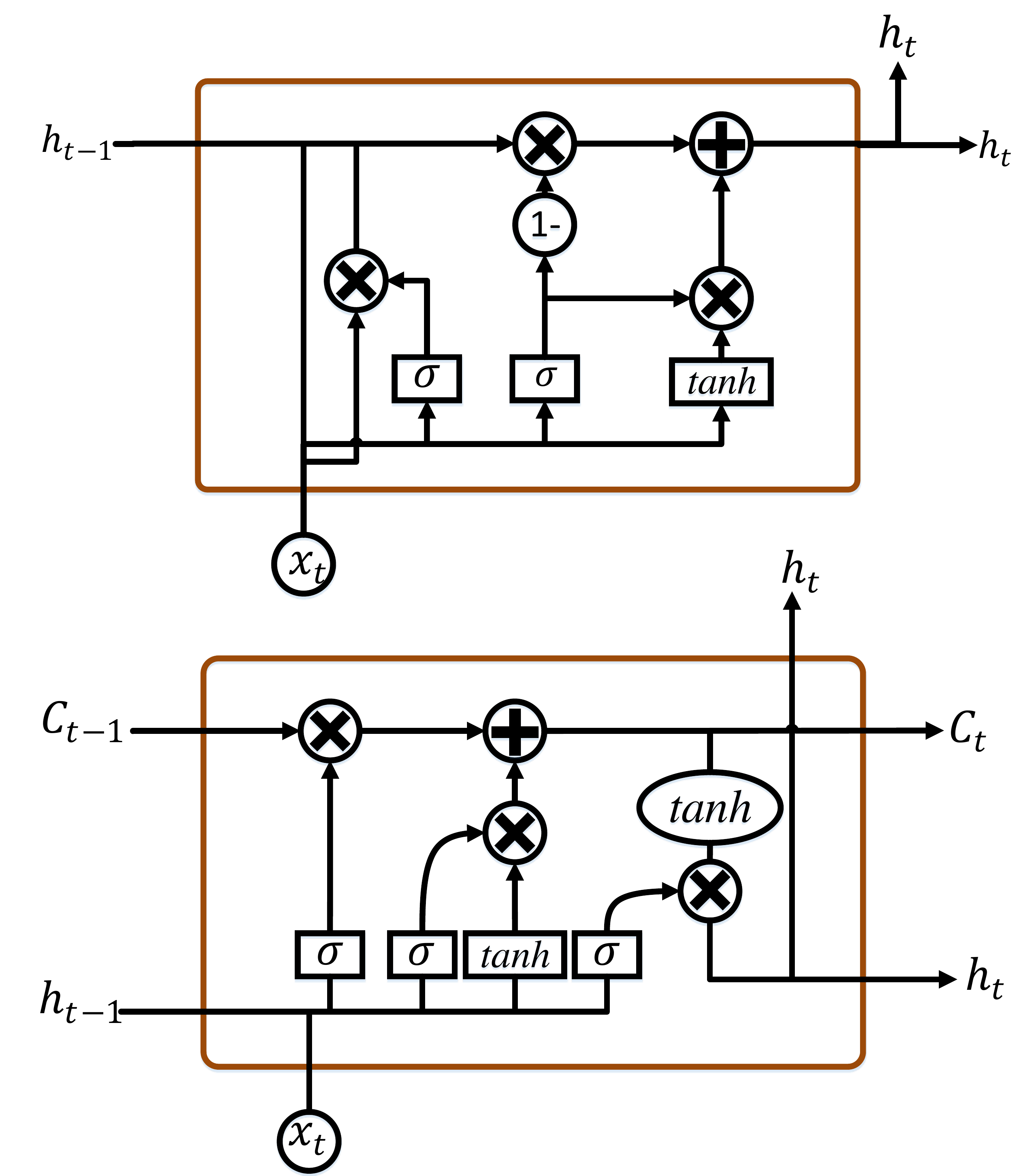}
\vspace{-0.15in}
\caption{Top Figure is a cell of GRU, and bottom Figure is a cell of LSTM}\label{fig:LSTM}

\end{figure}

Again, we have modified the basic architecture for use RMDL. Figure~\ref{Fig_RMDL} left side shows this extended RNN architecture. Several problems arise from RNN when the error of the gradient descent algorithm is back propagated through the network: vanishing gradient and exploding gradient ~\cite{bengio1994learning}. \\
\textbf{Long Short-Term Memory~(LSTM)}: To deal with these problems Long Short-Term Memory~(LSTM) is a special type of RNN that preserve long term dependency in a more effective way in comparison to the basic RNN. This is particularly useful to overcome vanishing gradient problem~\cite{pascanu2013difficulty}. Although LSTM has a chain-like structure similar to RNN,
LSTM uses multiple gates to carefully regulate the amount of information that will be allowed into each node state. Figure~\ref{fig:LSTM} shows the basic cell of a LSTM model. A step by step explanation of a LSTM cell is as following:

\begin{align}
    &&i_{t}=&\sigma(W_{i}[x_{t},h_{t-1}]+b_{i}),&& \label{eq:lstm1}\\
    &&\tilde{C_{t}}=&\tanh(W_{c}[x_{t},h_{t-1}]+b_{c}),&& \label{eq:lstm2} \\
    &&f_{t}=&\sigma(W_{f}[x_{t},h_{t-1}]+b_{f}),&& \label{eq:lstm3}\\
    &&C_{t}=&  i_{t}* \tilde{C_{t}}+f_{t} C_{t-1},&& \label{eq:lstm4}\\
    &&o_{t}=& \sigma(W_{o}[x_{t},h_{t-1}]+b_{o}),&& \label{eq:lstm5}\\
    &&h_{t}=&o_{t}\tanh(C_{t}),&&\label{eq:lstm6}
\end{align}
Where equation~\ref{eq:lstm1} is input gate, Equation~\ref{eq:lstm2} shows candid memory cell value, Equation~\ref{eq:lstm3} is forget gate activation, Equation~\ref{eq:lstm4} is new memory cell value, and  Equation~\ref{eq:lstm5} and~\ref{eq:lstm6} show output gate value. In the above description all $b$ represents bias vectors and all $W$ represent weight matrices and $x_{t}$ is used as input to the memory cell at time~$t$. Also,~$i,c,f,o$ indices refer to input, cell memory, forget and output gates respectively.
 Figure~\ref{fig:LSTM} shows the structure of these gates with a graphical representation.\\
An RNN can be biased when later words are more influential than the earlier ones. To overcome this bias Convolutional Neural Network~(CNN)  models~(discussed in Subsection~\ref{subsubsec:CNN} were introduced which deploys a max-pooling layer to determine discriminative phrases in a text~\cite{lai2015recurrent}.\\

\textbf{Gated Recurrent Unit~(GRU):}\label{subsec:GRU}
Gated Recurrent Unit~(GRU) is a gating mechanism for RNN which was introduced by~\cite{chung2014empirical} and~\cite{cho2014learning}. GRU is a simplified variant of the
LSTM architecture, but there are differences as follows: GRU contains two gates, a GRU does not possess internal memory (the $C_{t-1}$ in Figure~\ref{fig:LSTM}); and finally, a second non-linearity is not applied~(tanh in Figure~\ref{fig:LSTM}). A step by step explanation of a GRU cell is as following:
\begin{equation}
z_{t}=\sigma_g(W_{z}x_{t}+U_zh_{t-1}+b_{z}), \label{eq:gru1}
\end{equation}
Where~$z_t$ refers to update gate vector of~$t$,~$x_t$ stands for input vector,~$W$, $U$ and~$b$ are parameter matrices and vector, $\sigma_g$ is activation function that could be sigmoid or ReLU.
\begin{equation}
    \tilde{r_{t}}=\sigma_g(W_{r}x_{t}+U_rh_{t-1}+b_{r}), \label{eq:gru2}
\end{equation}
\begin{equation}
   h_t =  z_t \circ h_{t-1} + (1-z_t) \circ \sigma_h(W_{h} x_t + U_{h} (r_t \circ h_{t-1}) + b_h)\label{eq:gru6}
\end{equation}
Where~$h_t$ is output vector of~$t$, $r_t$ stands for reset gate vector of~$t$, $z_t$ is update gate vector of~$t$, $\sigma_h$ indicates the hyperbolic tangent function.

\subsubsection{Convolutional Neural Networks~(CNN)}\label{subsubsec:CNN}

The final deep learning approach which contributes in RMDL is  Convolutional Neural Networks~(CNN) that is employed for  document or image classification. Although originally built for image processing with architecture similar to the visual cortex, CNN have also been effectively used for  text classification ~\cite{lecun1998gradient}; thus, in RMDL, this technique is used in all datasets. \\ In the basic CNN for image processing an image tensor is convolved with a set of kernels of size $d \times d$. These convolution layers are called feature maps and can be stacked to provide multiple filters on the input. To reduce the computational complexity CNN use pooling which reduces the size of the output from one layer to the next in the network. Different pooling techniques are used to reduce outputs while preserving important features ~\cite{scherer2010evaluation}. The most common pooling method is max pooling where the maximum element is selected in the pooling window.\\ In order to feed the pooled output from stacked featured maps to the final layer, the maps are flattened into one column. The final layers in a CNN are typically fully connected.\\
In general, during the back propagation step of a convolutional neural network not only the weights are adjusted but also the feature detector filters. A potential problem of CNN used for text is the number of 'channels', $\Sigma$~(size of the feature space). This might be very large~(\textit{e.g.} 50K), for text but for images this is less of a problem~(\textit{e.g.} only 3 channels of RGB)~\cite{johnson2014effective}. This means the dimensionality of the CNN for text is very high.

\begin{figure}[t]
\centering

\includegraphics[width=0.6\columnwidth]{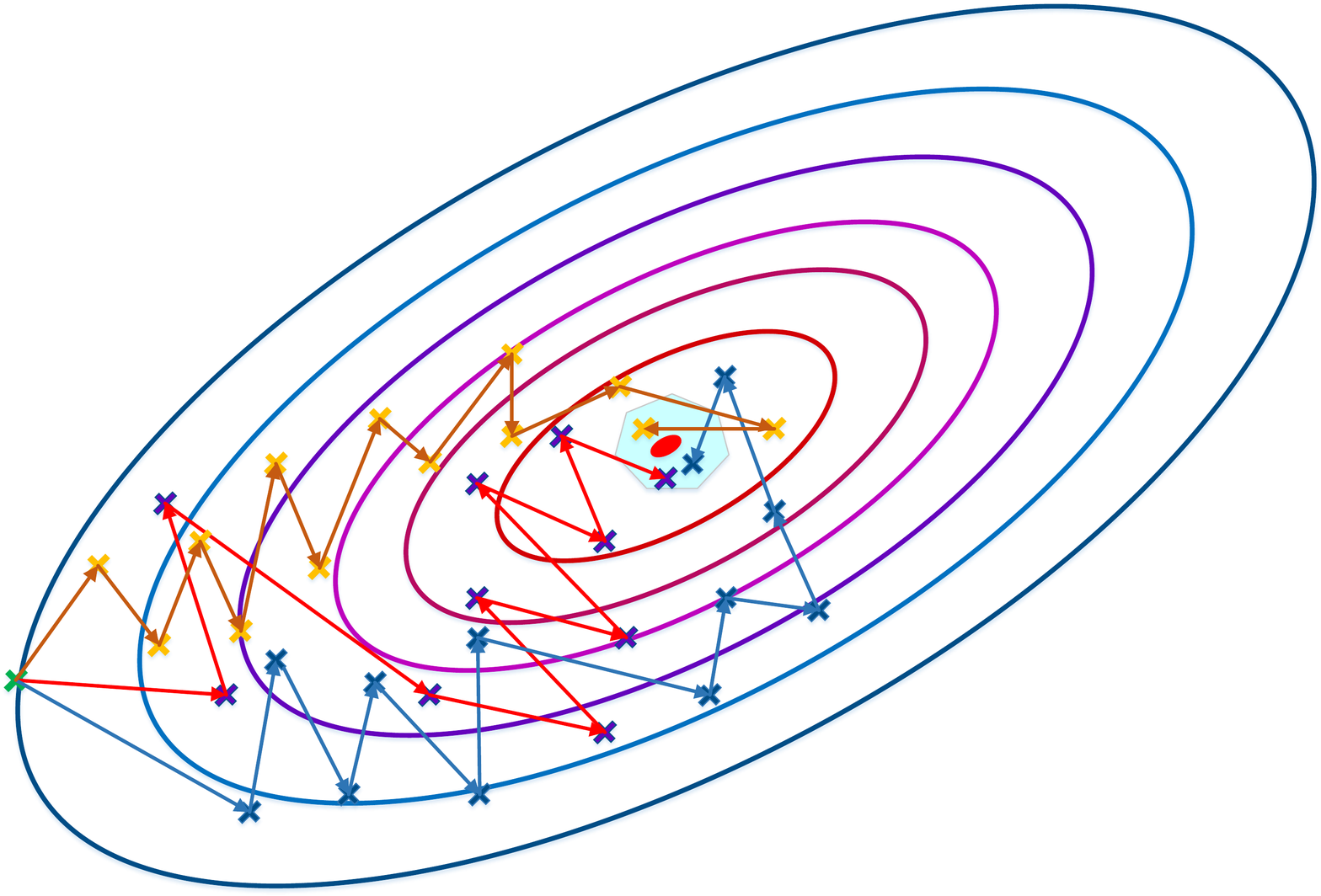}
\caption{This figure Shows multi SGD optimizer }\label{Optimizer}
\vspace{-0.15in}
\end{figure}

\subsection{Optimization}\label{subsec:Optimization}
In this paper we use two types of stochastic gradient optimizer in our neural networks implementation which are RMSProp  and Adam optimizer:
\subsubsection{Stochastic Gradient Descent~(SGD) Optimizer}
SGD has been used as one of our optimizer that is shown in equation~\ref{SGD}. It uses a momentum on re-scaled gradient which is shown in equation~\ref{momentum} for updating parameters. The other technique of optimizer that is used is RMSProp which does not do bias correction. This will be a significant problem while dealing with sparse gradient. 
\begin{align}
\label{SGD}
\theta &\leftarrow \theta - \alpha \nabla_\theta J(\theta , x_i,y_i)\\
\label{momentum}
\theta &\leftarrow \theta -\big( \gamma \theta + \alpha \nabla_\theta J(\theta , x_i,y_i)\big)
\end{align}
\subsubsection{Adam Optimizer}
Adam is another stochastic gradient optimizer which uses only the first two moments of gradient~($v$ and $m$ that are shown in equation~\ref{adam},   \ref{adam1}, \ref{adam2}, and \ref{adam3}) and average over them. It can handle non-stationary of objective function as in RMSProp while overcoming the sparse gradient issue that was a drawback in RMSProp~\cite{kingma2014adam}.
\begin{align}
\label{adam}
\theta & \leftarrow \theta - \frac{\alpha}{\sqrt{\hat{v}}+\epsilon} \hat{m}\\
g_{i,t} &=  \nabla_\theta J(\theta_i , x_i,y_i) \label{adam1}\\
\label{adam2}
m_t &= \beta_1 m_{t-1} + (1-\beta_1)g_{i,t}\\
\label{adam3}
m_t &= \beta_2 v_{t-1} + (1-\beta_2)g_{i,t}^2
\end{align}
Where $m_t$ is the first moment and $v_t$ indicates second moment that both are estimated. $\hat{m_t}=\frac{m_t}{1-\beta_1^t}$ and $\hat{v_t}=\frac{v_t}{1-\beta_2^t}$

\subsubsection{Multi Optimization rule}
The main idea of using multi model with different optimizers is that if one optimizer does not provide a good fit for a specific datasets, the RMDL model with~$n$ random models~(some of them might use different optimizers) could ignore~$k$ models which are not efficient if and only if $n>k$. The Figure~\ref{Optimizer} provides a visual insight on how three optimizers work better in the concept of majority voting. Using multi techniques of optimizers such as SGD, adam, RMSProp, Adagrad, Adamax, and so on helps the RMDL model to be more stable for any type of datasets. In this research, we only used two optimizers~(Adam and RMSProp) for evaluating our model, but the RMDL model has the capability to use any kind of optimizer.

\section{Experimental Results}\label{sec:results}
\begin{figure*}[htbp]
    \centering
    \begin{subfigure}[t]{0.479\textwidth}
        \includegraphics[width=\textwidth]{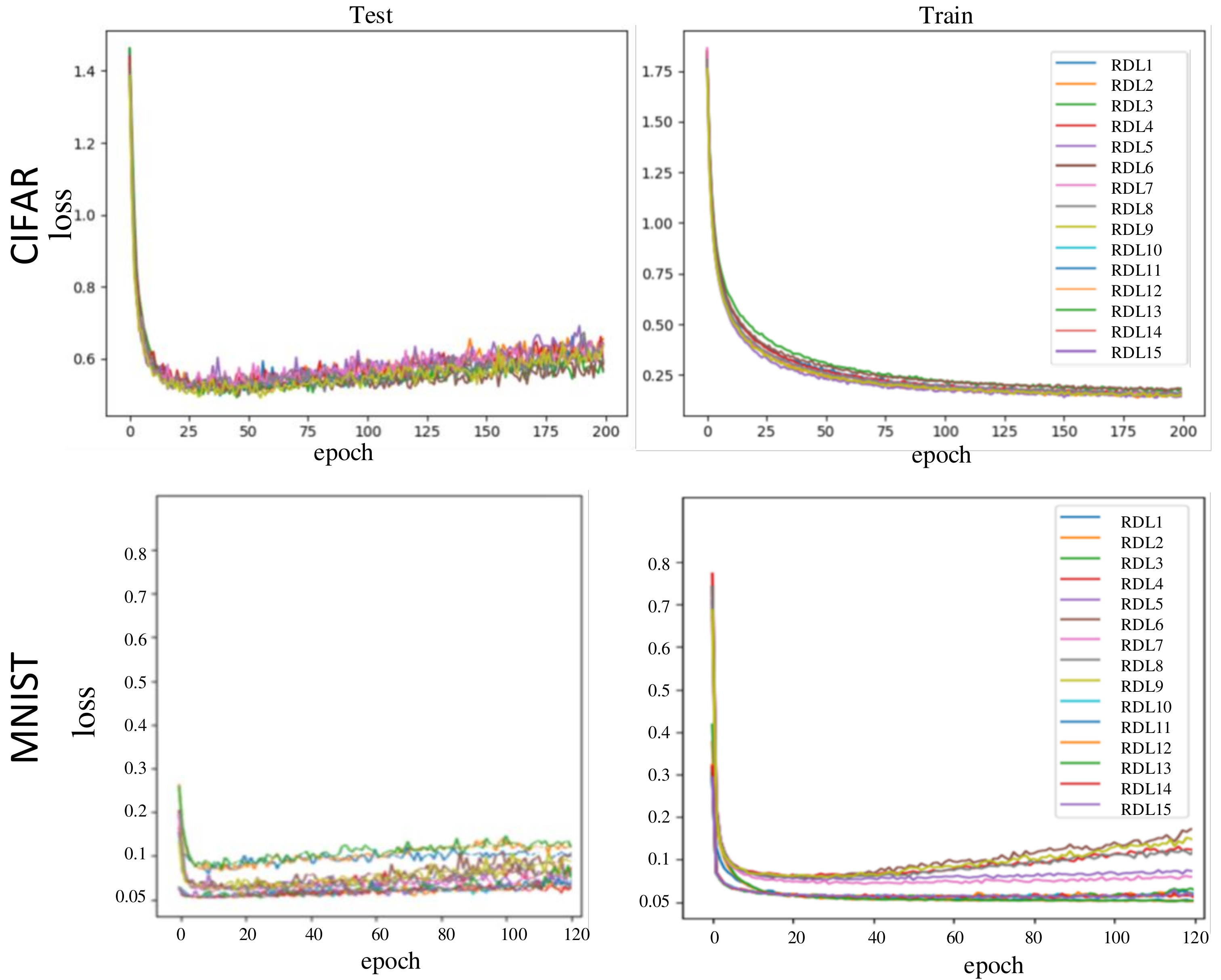}
        \caption{This sub-figure indicates MNIST and CIFAR-10 loss function for 15 Random Deep Learning~(RDL) model. The MNNST shown as 120 epoch and CIFAR has 200 epoch}\label{fig:image_loss}
    \end{subfigure}
  \hfill
    \begin{subfigure}[t]{0.5\textwidth}
        \includegraphics[width=\textwidth]{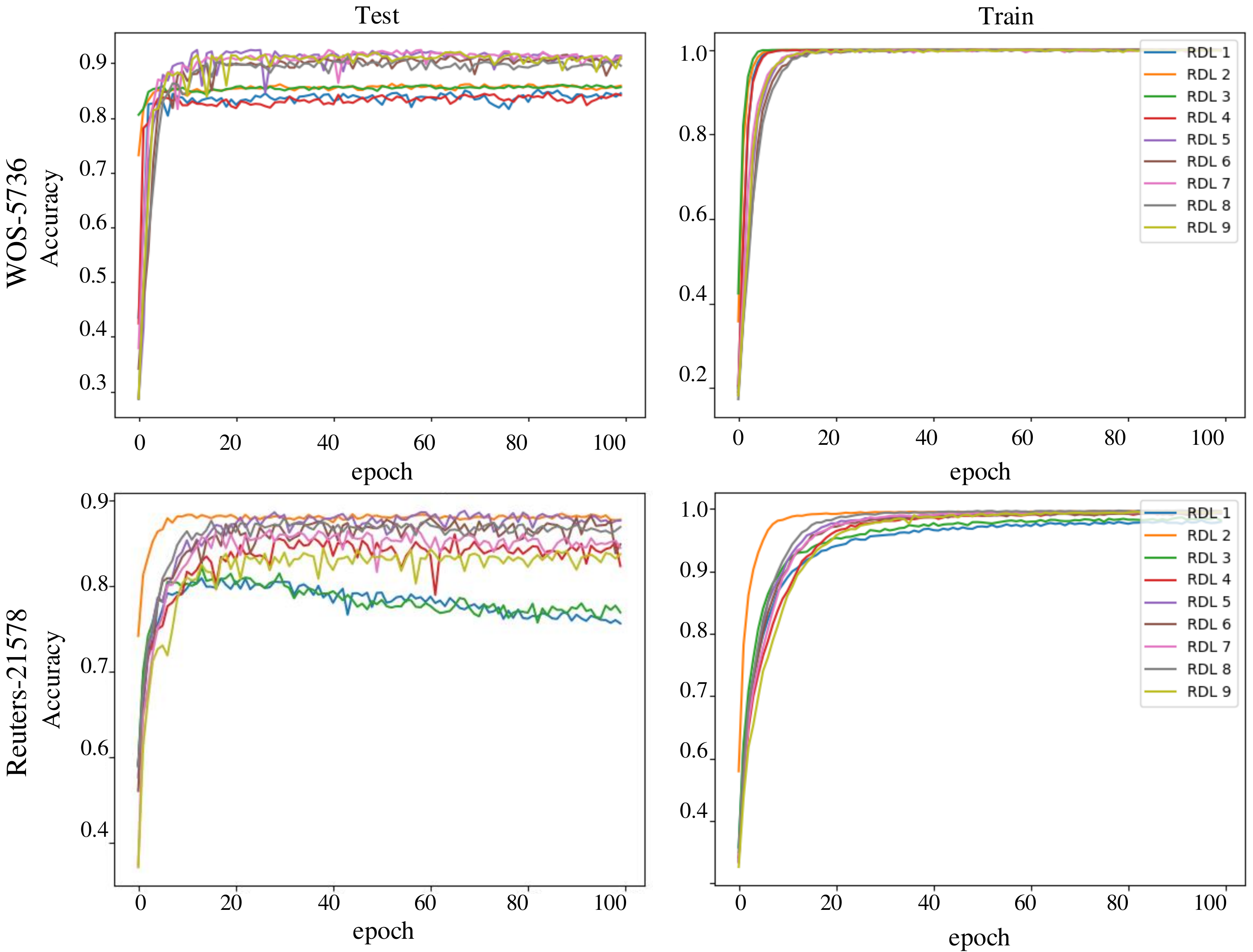}
        \caption{This sub-figure indicates WOS-5736~(\textit{Web Of Science dataset} with 11 categories and 5736 documents) accuracy function for 9 Random Deep Learning~(RDL) model, and bottom figure indicates Reuters-21578 accuracy function for 9 Random Deep Learning~(RDL) model}
        \label{fig:text_accuracy}
    \end{subfigure}
    \caption{This figure shows results of individual RDLs~(accuracy and loss) for each epoch as part of RMDL.}
    \label{fig:results}
 
\end{figure*}

In this section, experimental results are discussed including evaluation of method, experimental setup, and datasets. Also, we discuss the hardware and frameworks which are used in RMDL; finally, a comparison between our empirical results and the baselines has been presented. Moreover, losses and accuracies of this model for each individual RDL~(in each epoch) is shown in Figure~\ref{fig:results}.

\subsection{Evaluation}\label{subsec:Evaluation}
In this work, we report accuracy and \textit{Micro F1-Score} which are given as follows:
\begin{align}
Precision_{micro} &= \frac{\sum_{l=1}^LTP_l}{\sum_{l=1}^LTP_l+FP_l}\\
Recall_{micro} &= \frac{\sum_{l=1}^LTP_l}{\sum_{l=1}^LTP_l+FN_l}\\
F1-Score_{micro} &=  \frac{\sum_{l=1}^L2TP_l}{\sum_{l=1}^L2TP_l+FP_l+FN_l}
\end{align}
However, the performance of our model is evaluated only in terms of F1-score for evaluation as in Tables~\ref{ta:text} and~\ref{ta:text2}.
Formally, given~$I = \{1,2,\cdots, k\}$ a set of indices, we define the~$i^{th}$ class as~$C_i$. If we denote~$l = |I|$ and for~$TP_i$-true positive of~$C_i$,~$FP_i$-false
positive,~$FN_i$-false negative, and~$TN_i$-true negative counts respectively then the above definitions apply for our multi-class classification problem.

\begin{table}[b]
\centering
\vspace{-0.15in}
\caption{Accuracy comparison for text classification. W.1~(WOS-5736) refers to Web of Science dataset, W.2 represents W-11967, W.3 is WOS-46985, and R stands for Reuters-21578}
\label{ta:text}
\vspace{-0.15in}
\begin{tabular}{|c | c l c l c l c l c l|}
\hline
\multirow{2}{*}{}         & \multicolumn{2}{c|}{\multirow{2}{*}{Model}} & \multicolumn{8}{c|}{Dataset}                                                                                                                          \\ \cline{4-11} 
                          & \multicolumn{2}{c|}{}                       & \multicolumn{2}{c}{W.1}          & \multicolumn{2}{c }{W.2}          & \multicolumn{2}{c}{W.3}          & \multicolumn{2}{c|}{R}  \\ \hline
\multirow{9}{*}{Baseline} & \multicolumn{2}{c|}{DNN}                    & \multicolumn{2}{c}{86.15}          & \multicolumn{2}{c}{80.02}          & \multicolumn{2}{c}{66.95}          & \multicolumn{2}{c|}{85.3}           \\ \cline{2-11} 
                          & \multicolumn{2}{c|}{CNN~\cite{yang2016hierarchical}}      & \multicolumn{2}{c}{88.68}          & \multicolumn{2}{c }{83.29}          & \multicolumn{2}{c}{70.46}          & \multicolumn{2}{c|}{86.3}           \\ \cline{2-11} 
                          & \multicolumn{2}{c|}{RNN~\cite{yang2016hierarchical}}       & \multicolumn{2}{c}{89.46}          & \multicolumn{2}{c }{83.96}          & \multicolumn{2}{c}{72.12}          & \multicolumn{2}{c|}{88.4}           \\ \cline{2-11} 
                          & \multicolumn{2}{c|}{NBC}                    & \multicolumn{2}{c}{78.14}          & \multicolumn{2}{c }{68.8}           & \multicolumn{2}{c}{46.2}           & \multicolumn{2}{c|}{83.6}           \\ \cline{2-11} 
                          & \multicolumn{2}{c|}{SVM~\cite{zhang2008text}}       & \multicolumn{2}{c}{85.54}          & \multicolumn{2}{c }{80.65}          & \multicolumn{2}{c}{67.56}          & \multicolumn{2}{c|}{86.9}           \\ \cline{2-11} 
                          & \multicolumn{2}{c|}{SVM~(TF-IDF)~\cite{chen2016turning}}        & \multicolumn{2}{c}{88.24}          & \multicolumn{2}{c }{83.16}          & \multicolumn{2}{c}{70.22}          & \multicolumn{2}{c|}{88.93}           \\ \cline{2-11} 
                          & \multicolumn{2}{c|}{Stacking SVM~\cite{sun2001hierarchical}}  & \multicolumn{2}{c}{85.68}          & \multicolumn{2}{c }{79.45}          & \multicolumn{2}{c}{71.81}          & \multicolumn{2}{c|}{NA}             \\ \cline{2-11}
                          & \multicolumn{2}{c|}{HDLTex~\cite{kowsari2017HDLTex}}  & \multicolumn{2}{c}{90.42}          & \multicolumn{2}{c }{86.07}          & \multicolumn{2}{c}{76.58}          & \multicolumn{2}{c|}{NA}             \\ \hline
\multirow{3}{*}{RMDL}     & \multicolumn{2}{c|}{3 RDLs}        & \multicolumn{2}{c}{\textbf{90.86}} & \multicolumn{2}{c}{\textbf{87.39}} & \multicolumn{2}{c}{\textbf{78.39}} & \multicolumn{2}{c|}{\textbf{89.10}} \\ \cline{2-11} 
                          & \multicolumn{2}{c|}{9 RDLs}        & \multicolumn{2}{c}{\textbf{92.60}} & \multicolumn{2}{c}{\textbf{90.65}} & \multicolumn{2}{c}{\textbf{81.92}} & \multicolumn{2}{c|}{\textbf{90.36}} \\ \cline{2-11} 
                          & \multicolumn{2}{c|}{15 RDLs}       & \multicolumn{2}{c}{\textbf{92.66}} & \multicolumn{2}{c}{\textbf{91.01}} & \multicolumn{2}{c}{\textbf{81.86}} & \multicolumn{2}{c|}{\textbf{89.91}} \\ \cline{2-11} 
                          
                          & \multicolumn{2}{c|}{30 RDLs}       & \multicolumn{2}{c}{\textbf{93.57}} & \multicolumn{2}{c}{\textbf{91.59}} & \multicolumn{2}{c}{\textbf{82.42}} & \multicolumn{2}{c|}{\textbf{90.69}} \\ \hline

\end{tabular}

\end{table}

\subsection{Experimental Setup} 
Two types of datasets~(text and image) has been used to test and evaluate our approach performance. However, in theory the model has capability to solve classification problems with a variety of data including video, text, and images.  
\subsubsection{Text Datasets}
For text classification, we used~$4$ different datasets, namely, $WOS$ , $Reuters$, $IMDB$, and $20newsgroups$.\\ \textit{Web Of Science (WOS)} dataset~\cite{kowsari2018WOS} is a collection of academic articles' abstracts which contains three corpora~(\textit{5736, 11967, and 46985 }documents) for (\textit{11, 34, and 134} topics).\\ The \textit{Reuters-21578}  news dataset contains $10,788$ documents which are divided into $7,769$ documents for training and $3,019$ for testing with total of $90$ classes.\\
\textit{IMDB} dataset contains~$50,000$ reviews that is splitted into a set of~$25,000$ highly popular movie reviews for training, and~$25,000$ for testing.\\
\textit{20NewsGroup} dataset includes~$19,997$ documents with maximum length of~$1,000$ words. In this dataset, we have~$15,997$ for training and~$4,000$ samples are used for validation.
\subsubsection{Image datasets}
For image classification, two traditional and ground truth datasets are used, namely, \textit{MNIST} hand writing dataset and \textit{CIFAR}.\\\textit{MNIST:} this dataset contains handwritten number $k\in\{0,1,...,9\}$ and input feature space is in~$28\times28\times1$ format. The training and the test set contains~$60,000$ and ~$10,000$ data point examples respectively.\\ \textit{CIFAR:} This dataset consists of~$60,000$ images with~$32\times32\times3$ format assigned in~$10$ classes, with~$6,000$ images per class that is splitted into~$50,000$ training and~$10,000$ test images. Classes are airplane, automobile, bird, cat, deer, dog, frog, horse, ship, and truck.

\subsection{Hardware}
All of the results shown in this paper are performed on Central Process Units~(CPU) and Graphical Process Units~(GPU). Also, RMDL can be implemented using only GPU, CPU, or both. The processing units that has been used through this experiment was intel on \textit{Xeon~E5-2640~ (2.6 GHz)} with~\textit{12 cores} and \textit{64~GB} memory~(DDR3). Also, we have used three graphical cards on our machine which are two~\textit{Nvidia GeForce GTX~1080~Ti} and \textit{Nvidia Tesla~K20c}.
\subsection{Framework}
This work is implemented in Python using Compute Unified Device Architecture~(CUDA) which is a parallel computing platform and Application Programming Interface ~(API) model created by $Nvidia$. We used $TensorFelow$ and $Keras$ library for creating the neural networks~\cite{abadi2016tensorflow,chollet2015keras}. 

\begin{table}[t]
\centering
\caption{Error rate comparison for Image classification~(MNIST and CIFAR-10 datasets)}
\label{ta:image}
\vspace{-0.1in}
\begin{tabular}{|c | c | c c|}
\hline
\multicolumn{2}{|c}{Methods}                                            & MNIST         & CIFAR-10     \\ \hline
\multirow{5}{*}{Baseline} & Deep L2-SVM~\cite{tang2013deep}                 & 0.87          & 11.9         \\ \cline{2-4} 
                          & Maxout Network~\cite{goodfellow2013maxout} & 0.94          & 11.68        \\ \cline{2-4} 
                          & BinaryConnect~\cite{courbariaux2015binaryconnect} & 1.29          & 9.90         \\ \cline{2-4} 
                          & PCANet-1 ~\cite{chan2015pcanet}                      & 0.62          & 21.33        \\ \cline{2-4} 
                          & gcForest~\cite{zhou2017deep}                    & 0.74          & 31.00        \\ \hline
\multirow{4}{*}{RMDL}     & 3 RDLs                              & \textbf{0.51}     & \textbf{9.89}    \\ \cline{2-4} 
                          & 9 RDLs                              & \textbf{0.41} & \textbf{9.1} \\ \cline{2-4} 
                          & 15 RDLs                             & \textbf{0.21} & \textbf{8.74}    \\ \cline{2-4}
                          \cline{2-4} 
                          & 30 RDLs                             & \textbf{0.18} & \textbf{8.79}    \\ \hline
\end{tabular}
\vspace{-0.1in}
\end{table}

\subsection{Empirical Results}\label{sec:Empirical_Results}
\subsubsection{Image classification}
Table~\ref{ta:image} shows the error rate of RMDL for image classification. The comparison between the RMDL with baselines~(as described in Section~\ref{subsec:baseline_image}), shows that the error rate of the RMDL for MNIST dataset has been improved to~$0.51$,~$0.41$, and~$0.21$ for $3$,~$9$ and~$15$ random models respectively. For the CIFAR-10 datasets, the error rate has been decreased for RMDL to~$9.89$,~$9.1$, $8.74$, and~$8.79$,using~$3$,~$9$,~$15$, and $30$ RDL respectively.

\subsubsection{Document categorization}
Table~\ref{ta:text} shows that for four ground truth datasets, RMDL improved the accuracy in comparison to the baselines. In Table~\ref{ta:text}, we evaluated our empirical results by four different RMDL models~(using~$3$,~$9$,~$15$, and~$30$ RDLs). For \textit{Web of Science~(WOS-5,736)} the accuracy is improved to~$90.86$,~$92.60$,~$92.66$, and $93.57$ respectively. For  \textit{Web of Science~(WOS-11,967)}, the accuracy is increased to~$87.39$,~$90.65$,~$91.01$, and~$91.59$ respectively, and for \textit{Web of Science~(WOS-46,985)} the accuracy has increased to~$78.39$,~$81.92$,~$81.86$, and~$82.42$ respectively. The accuracy of \textit{Reuters-21578} is~$88.95$,~$90.29$,~$89.91$, and~$90.69$ respectively. We report results for other ground truth datasets such as \textit{Large Movie Review Dataset (IMDB)} and \textit{20NewsGroups}. As it is mentioned in Table~\ref{ta:text2}, for two ground truth datasets, RMDL improves the accuracy. In Table~\ref{ta:text2}, we evaluated our empirical results of two datasets~(\textit{IMDB reviewer and 20NewsGroups}).The accuracy of \textit{IMDB} dataset is~$89.91$,~$90.13$, and~$90.79$ for~$3$,~$9$, and~$15$ RDLs respectively, whereas the accuracy of DNN is~$88.55\%$, CNN~\cite{yang2016hierarchical} is~$87.44\%$, RNN~\cite{yang2016hierarchical} is~$88.59\%$, Na\"{i}ve Bayes Classifier is~$83.19\%$, SVM~\cite{zhang2008text} is~$87.97\%$, and SVM~\cite{chen2016turning} using TF-IDF is equal to~$88.45\%$. The accuracy of \textit{20NewsGroup} dataset is~$86.73\%$,~$87.62\%$, and~$87.91\%$ for~3,~9, and 15 random models respectively, whereas the accuracy of DNN is~$86.50\%$, CNN~\cite{yang2016hierarchical} is~$82.91\%$, RNN~\cite{yang2016hierarchical} is~$83.75\%$, Na\"{i}ve Bayes Classifier is~$81.67\%$, SVM~\cite{zhang2008text} is~$84.57\%$, and SVM~\cite{chen2016turning} using TF-IDF is equal to~$86.00\%$.\\

\begin{table}[b]
\centering
\vspace{-0.25in}
\caption{Accuracy comparison for text classification on IMDB and 20NewsGroup datasets }
\vspace{-0.15in}
\label{ta:text2}
\begin{tabular}{|c|c|l|clcl|}
\hline
\multirow{2}{*}{}     & \multicolumn{2}{c|}{\multirow{2}{*}{Model}}                 & \multicolumn{4}{c|}{Dataset}                                \\ \cline{4-7} 
                      & \multicolumn{2}{c|}{}                    & \multicolumn{2}{c}{IMDB}  & \multicolumn{2}{c|}{20NewsGroup} \\ \hline
\multirow{5}{*}{Baseline}             & \multicolumn{2}{c|}{DNN}                                    & \multicolumn{2}{c}{88.55} & \multicolumn{2}{c|}{86.50}            \\ \cline{2-7} 
                      & \multicolumn{2}{c|}{CNN~\cite{yang2016hierarchical}}          & \multicolumn{2}{c}{87.44} & \multicolumn{2}{c|}{82.91}            \\ \cline{2-7} 
                      & \multicolumn{2}{c|}{RNN~\cite{yang2016hierarchical} }         & \multicolumn{2}{c}{88.59} & \multicolumn{2}{c|}{83.75}            \\ \cline{2-7} 
                      & \multicolumn{2}{c|}{Na\"{i}ve Bayes Classifier}                                    & \multicolumn{2}{c}{83.19}      & \multicolumn{2}{c|}{81.67}            \\ \cline{2-7} 
                      & \multicolumn{2}{c|}{SVM~\cite{zhang2008text}  }               & \multicolumn{2}{c}{87.97}      & \multicolumn{2}{c|}{84.57}            \\ \cline{2-7} 
                      & \multicolumn{2}{c|}{SVM(TF-IDF)~\cite{chen2016turning}} & \multicolumn{2}{c}{88.45}      & \multicolumn{2}{c|}{86.00}            \\ \hline
\multirow{3}{*}{RMDL} & \multicolumn{2}{c|}{3 RDLs}                        & \multicolumn{2}{c}{\textbf{89.91}} & \multicolumn{2}{c|}{\textbf{86.73}}            \\ \cline{2-7} 
                      & \multicolumn{2}{c|}{9 RDLs}                        & \multicolumn{2}{c}{\textbf{90.13}} & \multicolumn{2}{c|}{\textbf{87.62}}            \\ \cline{2-7} 
                      & \multicolumn{2}{c|}{15 RDLs}                       & \multicolumn{2}{c}{\textbf{90.79}} & \multicolumn{2}{c|}{\textbf{87.91}}            \\ \hline
\end{tabular}

\end{table}

Figure~\ref{fig:results} indicates accuracies and losses of RMDL which are shown with~$9$~(RDLs) for text classification and~$15$ RDLs for image classification. As shown in Figure~\ref{fig:image_loss},~$4$ RDLs' loss of MNIST dataset are increasing over each epoch~(RDL~$6$, RDL~$9$, RDL~$14$ and RDL~$15$) after~$40$ epochs, but RMDL model contains~$15$ RDL models; thus, the accuracy of the majority votes for these models as presented in Table~\ref{ta:image} is competing with our baselines.\\ In Figure~\ref{fig:image_loss}, for CIFAR dataset, the models do not have overfitting problem, but for MNIST datasets at least 4 models' losses are increasing over each epoch after~$40$ iterations~(RDL~$4$, RDL~$5$, RDL~$6$, and RDL~$9$); although the accuracy and F1-measure of these~$4$ models will drop after~$40$ epochs, the majority votes' accuracy is robust and efficient which means RMDL will ignore them due to majority votes between~$15$ models. The Figure~\ref{fig:image_loss} shows the loss value over each epoch of two ground truth datasets, \textit{CIFAR} and \textit{IMDB} for~$15$ random deep learning models~(RDL). Figure~\ref{fig:text_accuracy} presents the accuracy of~15 random models for \textit{Reuters-21578} respectively. In Figure~\ref{fig:text_accuracy}, the accuracy of Random Deep Learning~(RDLs) model is addressed over each epoch for  WOS-5736~(\textit{Web Of Science dataset} with 17 categories and~$5,736$ documents), the majority votes of these models as shown in Table~\ref{ta:text} is competing with our baselines.

\vspace{-0.1in}
\section{Discussion  and Conclusion}\label{sec:Conclusion}

The classification task is an important problem to address in machine learning, given the growing number and size of datasets that need sophisticated classification.  We propose a novel technique to solve the problem of choosing best technique and method out of many possible structures and architectures in deep learning. This paper introduces a new approach called RMDL~(Random Multimodel Deep Learning) for the classification that combines multi deep learning approaches to produce 
random classification models. Our evaluation on datasets obtained from the Web of Science~(WOS), Reuters, MNIST, CIFAR, IMDB, and 20NewsGroups shows that combinations of DNNs, RNNs and CNNs with the parallel learning architecture, has consistently higher accuracy than those obtained by conventional approaches using na\"{i}ve Bayes, SVM, or single deep learning model. These results show that deep learning methods can provide improvements for classification and that they provide flexibility to classify datasets by using majority vote. The proposed approach has the ability to improve accuracy and efficiency of models and can be use across a wide range of data types and applications.




\bibliographystyle{ACM-Reference-Format}
\bibliography{ref}

\end{document}